\documentclass[letterpaper, 10pt, conference]{ieeeconf}  

\IEEEoverridecommandlockouts                              
\usepackage{graphics}
\usepackage{epsfig}
\usepackage{lipsum}
\usepackage{mathptmx}
\usepackage{times}
\usepackage{amsmath}
\usepackage{amssymb}
\usepackage{acronym}
\usepackage{graphicx}
\usepackage{multirow}
\usepackage{booktabs}
\usepackage{arydshln}

\usepackage[style=base]{caption}
\usepackage{subcaption}
\usepackage{xurl}
\usepackage{pifont}

    \newacro{cnn}[CNN]{Convolutional Neural Network}
    \newacro{jaad}[JAAD]{Joint Attention in Autonomous Driving}
    \newacro{adas}[ADAS]{Advanced Driver-Assistance Systems}
    \newacro{pie}[PIE]{Pedestrian Intention Estimation}
    \newacro{lstm}[LSTM]{Long Short-Term Memory}
    \newacro{bdlstm}[BDLSTM]{Bidirectional Long Short Term Memory}
    \newacro{bdgru}[BDGRU]{Bidirectional Gated Recurrent Unit}
    \newacro{seq2seq}[Seq2Seq]{sequence-to-sequence}
    \newacro{nlp}[NLP]{Natural Language Processing}
    \newacro{who}[WHO]{World Health Organization}
    \newacro{eu}[EU]{European Union}
    \newacro{etsc}[ETSC]{European Transport Safety Council}
    \newacro{iv}[IV]{Intelligent Vehicles}
    \newacro{vru}[VRU]{Vulnerable Road User}
    \newacro{ap}[AP]{Average Precision}
    \newacro{gru}[GRU]{Gated Recurrent Units}
    \newacro{gru}[GRU]{Gated Recurrent Unit}
    \newacro{rnn}[RNN]{Recurrent Neural Network}
    \newacro{bce}[BCE]{Binary Cross Entropy}
    \newacro{sota}[SOTA]{state of the art}
\title{\LARGE \bf
IntFormer: Predicting pedestrian intention with the aid of the Transformer architecture
}

\author{J. Lorenzo$^{1}$, I. Parra$^{1}$ and M. A. Sotelo$^{1}$
\thanks{
$^{1}$ Department of Computer Engineering, Universidad de Alcalá, Madrid, Spain
        {\{javier.lorenzod, ignacio.parra, miguel.sotelo\}@uah.es}
}
}

\begin{document}

\maketitle

\thispagestyle{empty}
\pagestyle{empty}

\begin{abstract}

Understanding pedestrian crossing behavior is an essential goal in intelligent vehicle development, leading to an improvement in their security and traffic flow. In this paper, we developed a method called IntFormer. It is based on transformer architecture and a novel convolutional video classification model called RubiksNet. Following the evaluation procedure in a recent benchmark, we show that our model reaches state-of-the-art results with good performance ($\approx 40$ seq. per second) and size ($8\times $smaller than the best performing model), making it suitable for real-time usage. We also explore each of the input features, finding that ego-vehicle speed is the most important variable, possibly due to the similarity in crossing cases in \ac{pie} dataset.

\end{abstract}
\section{Introduction}
\label{sec:intro}
Since the beginning of this century, several efforts have been made to improve road safety. According to data provided by the \ac{who} \cite{world_health_organization_global_2018}, the rate of road traffic deaths relative to world's population has stabilized and declined relative to the number of vehicles, but the progress is not enough to achieve a $50\%$ reduction by 2020 (SDG target 3.6) \cite{global_status_report_new}.

In \ac{eu}, pedestrians are the most affected \acp{vru} group in urban roads, representing nearly the $40\%$ of deaths, as stated by \ac{etsc} \cite{Adminaite-Fodor2019}. For this reason, urban scenarios driving, one of the major challenges that \ac{iv} are facing nowadays, must be tackled. In addition to obstacle detection and other \ac{adas}, the behavior anticipation of dynamic traffic participants  focused on the crossing action can lead to a risk reduction and a traffic flow improvement.

Over the last decade, human action recognition field has advanced greatly thanks to deep learning (read \cite{Zhu2020} for detailed information). Most of this methods, rely on hidden patterns found in camera sensor data to predict the correct action being performed by humans. Pedestrian crossing behavior anticipation can be viewed as an special case of action recognition, where the model, instead of predicting the action performed through the video, predicts the final action of the sequence with a limited observation interval as input. 

\begin{figure}[htbp]
    \centering
    \includegraphics[width=\columnwidth]{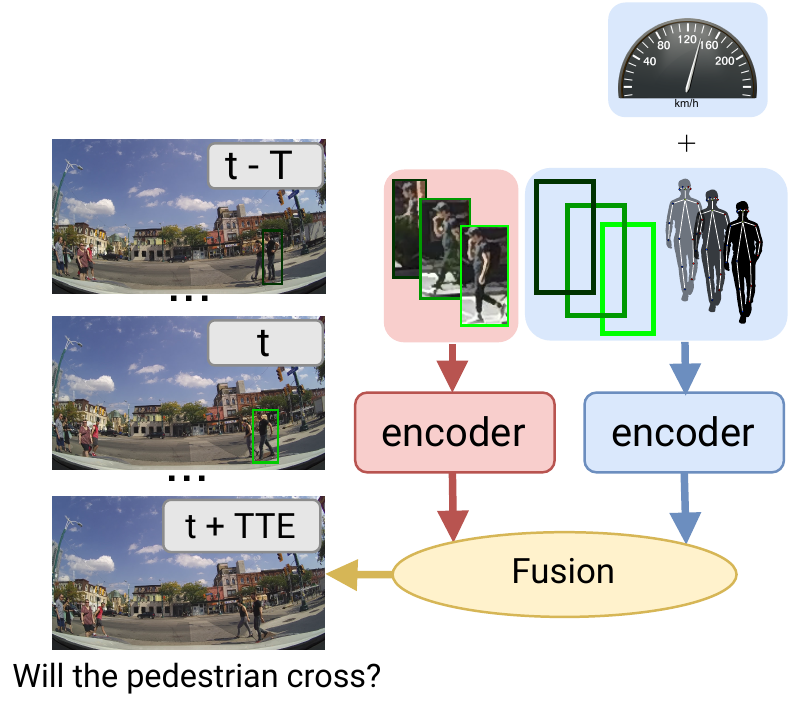}
    \caption{\small Diagram describing the proposed method. Data sequences of different nature (red means image data and blue non-image) are encoded and fused in order to estimate the final crossing behavior of the target pedestrian, which we refer to as intention.}
    \label{fig:visual_abstract}
\end{figure}

Since the publication of \ac{jaad} dataset \cite{Rasouli2018a_jaad}, several methods based on neural networks tried to deal with this challenging task by exploiting temporal information. However, without a common evaluation benchmark, the study of their performance was nearly impossible. Early this year, \ac{jaad} and \ac{pie} authors published a benchmark \cite{Kotseruba_benchmark} for both datasets. Several baselines were trained and evaluated and a state-of-the-art method called PCPA was proposed.

In this work, we propose a new method, IntFormer, which uses a state-of-the-art video action recognition model called RubiksNet for image bounding box sequences encoding and a transformer-based encoder for non-image data. It follows a similar architecture to the one proposed in PCPA model but trying to improve its performance while lowering its computational cost. To summarize, our contributions in this paper are:

\begin{itemize}
    \item A \textbf{novel composite model} for pedestrian action anticipation task. The proposed model is based on a novel video action recognition architecture, called \textbf{RubiksNet} \cite{fanbuch2020rubiks} and \textbf{transformer} architecture.
    \item State-of-the-art results with a \textbf{lighter model}, more suitable for \textbf{real-time applications}. 
    \item \textbf{Selection of input features} is key to better results: exploratory experiments are needed to understand the data used as input to the model.
    \item Different \textbf{preprocessing strategies} for input data used in the benchmark.
    \item \textbf{Faster training schedule}, by using novel optimization techniques.
\end{itemize}


\section{Related work}
\label{sec:related}
Two main approaches related to pedestrian crossing prediction. The first one is the human motion-based approach, where pedestrian behavior is implicitly extracted from the forecasted trajectory data (see \cite{Rudenko2020} for a detailed review of this methods).
The other branch, focuses on explicitly extracting pedestrian behavior from input data. The most extended source of information is video, and many of the recently published systems rely on techniques inherited from video action recognition field (see \cite{Zhu2020} for a detailed survey on this methods). In this field, the objective of the network is the estimation of the most suitable action class for an input video. However, this prediction is done for the current time, and in pedestrian crossing prediction the goal is on anticipation. In \cite{Saleh2019}, the authors
construct a complete pedestrian detection pipeline, with a 3D CNN-based network as the pedestrian crossing prediction system. In \cite{Piccoli2020}, a similar method is developed, which improves the previous one by including pose information extracted with a 2D CNN model. In \cite{Gujjar2019, Chaabane2020}, 3D CNN model is used for perform future frame prediction and crossing classification on that prediction. In \cite{Rasouli2019_pie, Rasouli2020b, Lorenzo2020, Kotseruba_benchmark}, a combination of 2D and recurrent based models are used, and in \cite{Kotseruba_benchmark} a 3D CNN backbone is used instead of a 2D one, due to its temporal advantages exploiting different sources of information rather than only image.

One of the main problems in this line of research is the lack of benchmarks, which does not allow a fair comparison between the different methods in the literature. As explained before, in \cite{Kotseruba_benchmark}, a benchmark and several baselines are provided based on \ac{jaad} dataset \cite{Rasouli2018a_jaad} and \ac{pie} dataset \cite{Rasouli2019_pie}. It is based on the problem of action anticipation rather than prediction as previous methods. Instead of predicting

\section{Method}
\label{sec:method}
\subsection{Problem formulation}
\label{subsec:problem-descr}
\subsubsection{Task description}

Pedestrian crossing action anticipation is defined as a binary classification task. Sequential information, recorded from the ego-vehicle, is used to predict the final crossing action of the target pedestrian track. Instead of using the whole length as in previous work, the last frame of observation $t$ is between $1$ and $2$ s ($30$ to $60$ frames) before the event (pedestrian starts to cross or the last frame where the pedestrian is observable in the case of non-crossing scenario). The input observation $N$ is $\approx 0.5$ s ($N = 16$ frames at $30$ fps). The prediction horizon varies as explained before, with a minimum of $1$ s and a maximum of $2$ s, which we will refer to as Time to event (TTE). For this reason, the time instant when pedestrian behavior is predicted is $t+TTE$.

\subsubsection{Input features}

In the proposed model, there are three main sources of information, all of them extracted from ground truth data available in the datasets or precomputed:
\begin{itemize}
    \item \textbf{Raw image}: Sequence of bounding boxes crops of pedestrians. In this case, we only use the \textit{local box} case of the benchmark, where only the image comprised inside the bounding box coordinates is used.
    \item \textbf{Bounding boxes image coordinates}: sequence of coordinates, to support raw image data providing localization of pedestrian in the image scene. 
    \item \textbf{Pose keypoints}: using the ones provided and used by PCPA model, obtained by using OpenPose \cite{Cao2016}. Being the output of a network, there are failure cases, where different limbs are not found on the pedestrian (zero values).
    \item \textbf{Ego-vehicle speed}: obtained from OBD sensor in PIE dataset and as a categorical value in JAAD. As it is not used in JAAD training of PCPA model, we do not use it either for a fairer comparison.
\end{itemize}

\subsection{Model architecture}

\begin{figure}[htb]
\centering
\includegraphics[width=0.9\columnwidth]{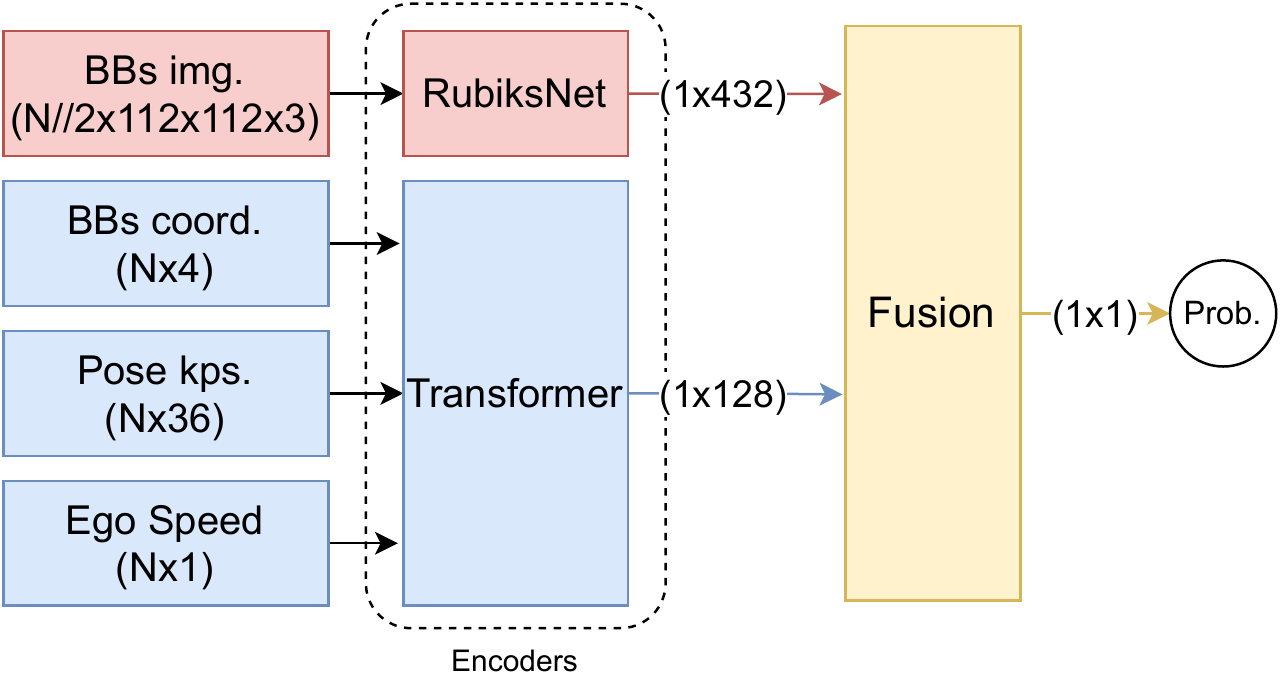}
  \caption{\small \textbf{Diagram of proposed model in detail}. Raw image sequence data of target pedestrian (delimited by bounding box coordinates) is used as input to RubiksNet \cite{fanbuch2020rubiks} backbone to obtain a feature vector. Bounding box coordinates, pose keypoints and ego-vehicle speed sequences are introduced to a transformer-based encoder which outputs a combined feature of these non-image features. Both outputs are introduced to the fusion block, concatenating them or applying many-to-one Luong attention.} 

\label{fig:pipeline}
\end{figure}

We present an end-to-end model, composed of three main parts:
\begin{itemize}
    \item \textbf{Raw video encoder}: composed by a pretrained spatio-temporal architecture. 3D CNN models have proven to be a better alternative to recurrent models in tasks where image data is used, such as action classification (see \cite{Zhu2020} for a detailed analysis of this technique). However, they are computationally expensive. Alternatives based on 2D convolutions with lower computational cost are also proposed in the literature \cite{Lin2018a}. In this model, temporal convolutions are replaced by channel-wise shifts operations. However 2D convolutions in this model remain the same. For this reason, in this block we planned to use RubiksNet \cite{fanbuch2020rubiks}, an efficient architecture for video action recognition that learns 3D-Shift operations and reduces the number of 2D convolutions, keeping a low number of parameters and FLOPS in comparison to other state-of-the-art video action recognition methods.
    
    \item \textbf{Non-image data encoder}: based on transformer architecture. This architecture was proposed in \cite{Vaswani2017AttentionIA} for NLP tasks, outperforming recurrent and convolutional models, which were state-of-the-art at that moment. Recently, several works have been published looking for ways to apply them to computer vision tasks \cite{Khan}.  This architecture relies on a self-attention mechanism, but, instead of using recurrence, transformer models create relationships between all samples in the input sequence, allowing parallelization and better use of modern devices such as TPUs and GPUs. In our case, we only keep the encoder part, using the output features as a representation. We combined all non-image sequence data as the input of a unique transformer encoder. After obtaining the combined output (fixed-size feature vector for each non-image sequence), we obtain the mean between all of them and apply a layer normalization and a fully connected layer.
    
    \item \textbf{Fusion module}: block in charge of merging features of the previous branches. It consists of a fully connected layer, whose input is composed of the concatenation of the outputs of previous encoders. An additional fully connected layer is used to obtain the final output logit. After the first fully connected layer, we apply a dropout of $0.5$ and a ReLU activation.
    
\end{itemize}

In the figure \ref{fig:pipeline} a detailed description of the model is shown.

\subsection{Data preprocessing}

In this section, we will discuss the operations performed on input data before using it as input to the network.

\subsubsection{Raw image data}: after the crop operation we obtain a sequence of $N$ images related to the target pedestrian. The benchmark established an input time length of $\approx 0.5$ s which corresponds to $16$ frames. In our experiments, We halved that number, sampling the input time interval at $15$ fps instead of $30$ fps, keeping sequences with $N / 2 = 8$ evenly spaced frames. Images are resized to square dimension $112\times 112\times 3$ ($H \times W \times C$ format). The input batch of the ResNet 3D model consists of a five-dimensional vector of shape $B \times C \times N / 2 \times H \times W$. In the case of RubiksNet, the input batch is a four-dimensional vector because it used bidimensional kernels. The shape is $B \times \frac{N \cdot C}{2} \times H\times W$.

No data augmentation is performed on the image. Min-max normalization is performed on input data for RubiksNet backbone.

\subsubsection{Bounding boxes coordinates}

In the benchmark, they used relative changes in bounding box coordinates concerning the first bounding box in the sequence. In our work, we perform a min-max normalization using image dimension ($1920\times 1080$ pixel).

\subsubsection{Pose keypoints}

Pose information is used the same way as in benchmark, performing a min-max normalization using the image dimension commented previously.

\subsubsection{Ego-vehicle speed}

In the benchmark, this variable is used without normalization. We performed a z-score normalization using the training set mean and standard deviation in all parts of the experiments (training, validation, and testing).

For the previous three non-image features, sequence length $N$ is used.

\begin{table*}[ht!]
\newcommand{\ra}[1]{\renewcommand{\arraystretch}{#1}}
\centering
\ra{1.2}
\resizebox{0.9\linewidth}{!}{%
\begin{tabular}{llllllllllll} 
\toprule
\multicolumn{1}{c}{\multirow{2}{*}{\textbf{Model }}} & \multicolumn{1}{c}{\multirow{2}{*}{\textbf{Fusion}}} & \multicolumn{1}{c}{\multirow{2}{*}{\textbf{\# P}}} & \multicolumn{3}{c}{\textbf{ PIE}} & \multicolumn{3}{c}{$\textrm{JAAD}_{\textrm{beh}}$} & \multicolumn{3}{c}{$\textrm{JAAD}_{\textrm{all}}$} \\ 
\cmidrule(lr){4-4}\cmidrule(lr){5-5}\cmidrule(lr){6-6}\cmidrule(lr){7-7}\cmidrule(lr){8-8}\cmidrule(lr){9-9}\cmidrule(lr){10-10}\cmidrule(lr){11-11}\cmidrule(lr){12-12}
\multicolumn{1}{c}{} & \multicolumn{1}{c}{} & \multicolumn{1}{c}{} & \textbf{Acc.} & \textbf{AUC} & \textbf{F1} & \textbf{Acc.} & \textbf{AUC} & \textbf{F1} & \textbf{Acc.} & \textbf{AUC} & \textbf{F1} \\ 
\cmidrule(lr){1-1}\cmidrule(lr){2-2}\cmidrule(lr){3-3}\cmidrule(r){4-6}\cmidrule(r){7-9}\cmidrule(lr){10-12}
PCPA\cite{Kotseruba_benchmark} & \multirow{2}{*}{A} & \multirow{2}{*}{31.2M} & 0.86 & 0.86 & 0.77 & 0.58 & 0.50 & \textbf{0.71} & 0.85 & \textbf{0.86} & \textbf{0.68} \\
PCPA (ours) &  &  & 0.86 & 0.86 & 0.78 & \multicolumn{1}{c}{-} & \multicolumn{1}{c}{-} & \multicolumn{1}{c}{-} & 0.84 & 0.74 & 0.56 \\ 
\hdashline
IntFormer & C & 4M & \textbf{0.89} & \textbf{0.92} & \textbf{0.81} & \textbf{0.59} & \textbf{0.54} & 0.69 & \textbf{0.86} & 0.78 & 0.62 \\
\bottomrule
\end{tabular}
}
\caption{\small Results in the benchmark. \# P means number of parameters. PCPA (Ours) is the model trained in our machine in order to reproduce PCPA results.}
\label{tab:results}
\end{table*}

\section{Evaluation}
\label{sec:exp_set}
\subsection{Data}

We have trained our models on both datasets, PIE, and JAAD. Inside JAAD, there are two variants, $\textrm{JAAD}_{\textrm{all}}$ and $\textrm{JAAD}_{\textrm{beh}}$. In the first case, it contains all annotated pedestrians and in the second, only pedestrians with behavioral annotations. This second case is the only dataset used in the benchmark with more crossing cases. Crossing class in PIE and complete JAAD are underrepresented.

\subsection{Training details}

All models in the performed experiments were trained using PyTorch framework through a PyTorch Lightning wrapper \cite{falcon2019pytorch}. During training, we have used different hyperparameters depending on the dataset:
\begin{itemize}
    \item \textbf{\ac{pie}}: batch size of $8$. After a small automatic hyperparameter search, we obtained the following parameters: RubiksNet learning rate of $1.1\cdot10^{-3}$, RubiksNet shift layers multiplier of $6.5\cdot 10^{-4}$ and transformer encoder's learning rate of $4.3\cdot 10^{-3}$. Using Adam optimizer with the default configuration of PyTorch.
    \item \textbf{$\textrm{JAAD}_{\textrm{beh}}$}: batch size of $8$. Unique learning rate of $10^{-4}$ ($10^{-5}$ for shift layers). Using AdamW \cite{loshchilov2019decoupled} for faster convergence and a weight decay of $10^{-3}$. In this experiment we obtained the results using only image and coordinates of the pedestrian bounding boxes. 
    \item \textbf{$\textrm{JAAD}_{\textrm{all}}$}: same training configuration than in the previous case, with a weight decay of $10^{-4}$ and unique learning rate of $3\cdot 10^{-4}$.
\end{itemize}

We have trained all our models on a Geforce GTX TITAN X (Pascal architecture) with a CPU i5-4690K \@ 3.50 GHz.

\subsubsection{Loss function}
We used binary cross-entropy loss. Because the benchmark is constructed over imbalanced data, we apply a weight $W_c$ for the positive class (crossing) to virtually equalize the weight of both classes. This weight calculation is shown in \ref{eq:weight_calculus}. $M_{nc}$ is the number of non-crossing cases. $M_c$ is the number of crossing cases. Both numbers are calculated in the training set. 
\begin{equation}
    W_c = M_{nc} / M_c
    \label{eq:weight_calculus}
\end{equation}

\subsection{Evaluation Metrics}

We use the same metrics as in the benchmark: accuracy, F1-Score, and area under the ROC curve (AUC).

\subsection{Baselines}

We compare our proposed architecture with the best performing one in the benchmark, called PCPA \cite{Kotseruba_benchmark}. This network is composed of a 3D convolutional network called C3D pretrained on Sports-1M dataset. This network is used as a video encoder and uses the full video ($N = 16$ frames). For non-image data, it uses \ac{gru} units and it uses two attention modules: one applied to the temporality and the other applied as the fusion strategy. Finally, it uses a fully connected layer to output the binary prediction.

\subsection{Input data importance}

In parallel with the benchmark, we experimented focused on the importance of the input data. Training on \ac{pie} dataset, we begin with the only video part of the model, and with each experiment, we include one feature. We also trained the model with one feature at a time, to see how each of them affects the performance of the model. For all cases, we used the same hyperparameters: AdamW optimizer with weight decay of $10^{-4}$ and unique learning rate of $3\cdot 10^{-4}$, and batch size of $8$ sequences.
\section{Results and Discussion}
\label{sec:results}
In table \ref{tab:results}, our proposed model is compared with PCPA. The best results in PIE dataset, which is the most diverse one, belongs to our model. It uses the same data, but the preprocessing is different (bounding boxes coordinates normalized instead of relative differences, image data at halved framerate, and normalized ego-vehicle speed). Instead of relying on an attention external block, our model performs a simple fusion by concatenating both groups of features (non-image and image data), which conforms the input of a fully connected layer. With respect to the model size, our model has a $13\%$ of PCPA parameter number and runs at $40$ sequences per second. Experiments with trained PCPA models (following official code) show a speed of $\approx 22.2 s$. However, the official code uses Tensorflow 2 instead of PyTorch, and the model occupies $11.479 $ GB of memory with a batch size of $1$. Our model, with the same tests, occupies less than $10\%$ ($1.2$ GB). With respect to the F1 Score, we observe that our model obtain a better value, but recall is better in PCPA (trained by ours), which means that the crossing cases are better covered. However, precision is nearly $8\%$ better in our model, which shows that it understands better the separation between the two cases.

With respect to \ac{jaad}, although we obtained similar results than PCPA with the behavioral data, our AUC result shows that our model performs better than the random case. In PCPA case, this value is near or equal to random performance ($AUC = 0.5$). 

Finally, results on PCPA using complete \ac{jaad} data show better results for PCPA model. However, after training PCPA with the official code on \ac{jaad}, we obtained worse results than our model. This could be due to the instability in training and the difficulty of the task. In both cases, the model performs better than in the random case.

Looking at the training time, our model takes advantage. Our training on \ac{pie} spends $\approx 8$ hours in PCPA ($80$ epochs). In the paper, they spent, assuming they only change the number of epochs, $6$ hours ($60$ epochs). In our case, $\approx 13 $min are used in training (less than a $4\%$). Training on \ac{jaad} follows a similar trend.

\subsection{Input data importance}

\begin{table}[h!]
\centering
\resizebox{\linewidth}{!}{%
\begin{tabular}{cccclll} 
\toprule
\textbf{Imgs} & \textbf{BBs } & \textbf{Pose} & \multicolumn{1}{l}{\textbf{\textbf{Speed}}} & \multicolumn{1}{c}{\textbf{Acc.}} & \textbf{AUC} & \textbf{F1} \\ 
\cmidrule(lr){1-1}\cmidrule(lr){2-2}\cmidrule(lr){3-3}\cmidrule(lr){4-4}\cmidrule(r){5-7}
\checkmark & - & - & - & 0.750 & 0.668 & 0.519 \\
- & \checkmark & - & - & 0.739 & 0.571 & 0.287 \\
- & - & \checkmark & - & 0.719 & 0.500 & 0.000 \\
- & - & - & \checkmark & \textbf{0.860} & 0.817 & 0.743 \\
- & - & \checkmark & \checkmark & 0.837 & 0.829 & 0.737 \\
- & \checkmark & - & \checkmark & 0.851 & \textbf{0.846} & \textbf{0.760} \\
\checkmark & - & - & \checkmark & 0.814 & 0.789 & 0.689 \\
- & \checkmark & \checkmark & - & 0.719 & 0.500 & 0.000 \\
\checkmark & - & \checkmark & - & 0.765 & 0.682 & 0.541 \\
- & \checkmark & \checkmark & \checkmark & 0.822 & 0.832 & 0.730 \\
\checkmark & \checkmark & \checkmark & - & 0.745 & 0.675 & 0.527 \\
\checkmark & - & \checkmark & \checkmark & 0.824 & 0.796 & 0.697 \\
\checkmark & \checkmark & - & \checkmark & 0.849 & 0.826 & 0.743 \\
\checkmark & \checkmark & - & - & 0.714 & 0.620 & 0.448 \\
\checkmark & \checkmark & \checkmark & \checkmark & 0.823 & 0.821 & 0.722 \\
\bottomrule
\end{tabular}
}
\caption{\small Results for the experiments carried out to analyze the importance of each input features. ``Imgs'' is bounding boxes image crops, ``BBs'' is bounding boxes coordinates, ``Pose'' is pose keypoints (precomputed) and ``Speed'' is ego-vehicle speed.}
\label{tab:input_data_importance}
\end{table}

In table \ref{tab:input_data_importance}, results for all combination of inputs is displayed. Individually, only pose information is irrelevant and the model behaves similarly to the random choice case ($AUC = 0.5$). This can be caused by the precomputed origin of this feature, which raises problems such as incorrect or missing detections and occlusions by other road participants. The second less informative variable is the bounding box coordinates. It is hand-labeled and obtains better individual results than pose. However, the lack of relation between samples can be the reason that leads to a worse temporal understanding by the transformer encoder. In the case of bounding boxes images, it is clear that this is a more informative variable than the previous ones, due to the inclusion of context on the image. Nevertheless, image data has large variability, and it is nearly impossible to generalize only with the available data. As the final and best feature, ego-vehicle speed achieves the best single performance, with a long advantage concerning the rest.

These results may mean that the crossing cases are closely related to the changes in speed, which could translate, for example, to a majority of crossing situations at zebra crossings, where pedestrians behave similarly. Also, this variable is unique for each sequence, so several pedestrians share the same speed.
Looking at the experiments using combined features it is clear that image bounding boxes benefit from non-image data processed with the transformer in a great way. The best case though, corresponds to the combination of bounding box coordinates and speed. Maybe this is caused by the low dimensionality of coordinates, which combined with the location information can lead to a better understanding of each pedestrian case, leveraging a distinction between cases. 

Pose information affects negatively in most of the cases, but the combination of features is not enough to assure that worse performance is caused by this feature since there are cases where it improves. 

Finally, the best performing model in this set relies only on the transformer model and achieves similar results to the ones of PCPA model in table \ref{tab:results}. 
\section{Conclusions}
\label{sec:conc_fut_work}
In this work, we have presented a novel architecture, called IntFormer, for the task of pedestrian crossing intention prediction. It is based on a novel video action recognition architecture called RubiksNet and transformer architecture. An exploratory study of the importance of variables has shown the big relation between ego-vehicle speed and the crossing intention, which can mean that most of the crossing cases in \ac{pie} dataset shares a similar pattern. Our model has been ranked on a recently published benchmark \cite{Kotseruba_benchmark}, achieving state-of-the-art results with less data, computational resources, and training time.

In the future, a deep analysis of datasets \ac{pie} and \ac{jaad} can be done to check the importance of each variable. Additional precomputed variables can be included to replace or complement the pose keypoints. Training pipeline optimization can be improved, by automatically looking for better hyperparameters.
\bibliographystyle{IEEEtran}
\bibliography{references}

\end{document}